\documentclass[journal,onecolumn,11pt]{IEEEtran}
\usepackage{caption}
\usepackage{graphicx}
 \graphicspath{{figures/}}
 \DeclareGraphicsExtensions{.pdf,.jpeg,.png,.jpg,}
 \usepackage{epstopdf}
 \usepackage[caption=false,font=footnotesize]{subfig}

\usepackage{amsmath}
\usepackage{amssymb}
\usepackage{bm}

\usepackage{fixltx2e}

\usepackage[space]{cite}
\usepackage{url}

\usepackage{tikz}
\usepackage{float}

\title{A Study of Deep Learning Robustness \\Against Computation Failures}
\author{
  Jean-Charles Vialatte and Fran\c{c}ois Leduc-Primeau
  \thanks{This is an extended version of the paper with the same title published in Proc. 9th Int.~Conf.~on Advanced Cognitive Technologies and Applications, Feb.~2017.}
  \thanks{Jean-Charles Vialatte and Fran\c{c}ois Leduc-Primeau are with IMT Atlantique, Brest, France. Jean-Charles Vialatte is also with Cityzen Data, Guipavas, France.}
  \thanks{Emails: \{jc.vialatte, francois.leduc-primeau\}@imt-atlantique.fr.}
}

\begin{document}
\maketitle
\pagestyle{empty} % No page numbers (must appear after \maketitle)
\thispagestyle{empty} % (both are required)

\begin{abstract}
% (François)
For many types of integrated circuits, accepting larger failure rates in computations can be used to improve energy efficiency.
We study the performance of faulty implementations of certain deep neural networks based on pessimistic and optimistic models of the effect of hardware faults.
After identifying the impact of hyperparameters such as the number of layers on robustness, we study the ability of the network to compensate for computational failures through an increase of the network size.
We show that some networks can achieve equivalent performance under faulty implementations, and quantify the required increase in computational complexity.
%This provides a bound on the energy savings that must be associated with the unreliable operation of the circuit.
\end{abstract}

%\begin{IEEEkeywords}
%Deep learning; quasi-synchronous circuits; energy-efficient computing.%
%\end{IEEEkeywords}

\section{Introduction}\label{sec:intro}
% (François)

Deep neural networks achieve excellent performance at various artificial intelligence tasks, such as speech recognition \cite{DBLP:journals/corr/AmodeiABCCCCCCD15} and computer vision \cite{krizhevsky:2012}.
Clearly, the usefulness of these algorithms would be increased significantly if state-of-the-art accuracy could also be obtained when implemented on embedded systems operating with reduced energy budgets.
In this context, the energy efficiency of the inference phase is the most important, since the learning phase can be performed offline.

To achieve the best energy efficiency, it is desirable to design specialized hardware for deep learning inference. However, whereas in the past the energy consumption of integrated circuits was decreasing steadily with each new integrated circuit technology, the energy improvements that can be expected from further shrinking of CMOS circuits is small \cite{dreslinski:2010}.
A possible approach to continue improving the energy efficiency of CMOS circuits is to operate them in the near-threshold regime, which unfortunately drastically increases the amount of delay variations in the circuit, which can lead to functional failures \cite{alioto:2012}.
There is therefore a conflict between the desire to obtain circuits that operate reliably and that are energy efficient. 
In other words, tolerating circuit faults without degrading performance translates into energy gains.
Neural networks are interesting algorithms to study in this context, because their ability to process noisy data could also be useful to compensate for hardware failures.
It is interesting to draw a parallel with another important class of algorithms that process noisy data: decoders of error-correction codes. For the case of low-density parity-check codes, it was indeed shown that a decoder can save energy by operating unreliably while preserving equivalent performance \cite{leduc-primeau:2015}.

Of course, a perhaps more straightforward way to decrease the energy consumption is to reduce the computational complexity. For example, \cite{iandola:2016} proposes an approach to decrease the number of parameters in a deep convolutional neural network (CNN) while preserving prediction accuracy, and \cite{courbariaux:2016,rastegari:2016} propose approaches to replace floating-point operations with much simpler binary operations.
The approach of this paper is the opposite. We consider increasing the number of parameters in the model to provide robustness that can then be traded for energy efficiency.
The two aims are most likely complementary, and the situation is in fact similar in essence to the problem of data transmission, where it is in many practical cases
%\footnote{For example, this was shown by Shannon for the case of point-to-point communication \cite{shannon:1948}.} 
asymptotically (in the size of the transmission) optimal to first compress the data to be transmitted, and then to add back some redundancy using an error-correction code.

In this preliminary study, we consider the ability of simple neural network models to increase their robustness to computation failures through an increase of the number of parameters.
It was shown previously for an implementation of a multilayer perceptron (MLP) based on stochastic computing that it is possible to compensate for a reduced precision by increasing the size of the network \cite{ardakani:2017}.
The impact of hardware faults on CNNs is also considered in \cite{lin:2016} using a different approach that consists in adding compensation mechanisms in hardware while keeping the same network size.
The deviation models that we consider in this paper attempt to represent the case in which computation circuits are not protected by any compensation mechanism. We consider a pessimistic model and a more optimistic deviation model, but in both cases the magnitude of the error is only limited by the bounded codomain of the computations. We show that despite this, increasing the size of the network can sometimes compensate for faulty computations. The results provide an idea of the reduction in energy that must be obtained by a faulty circuit in order to reduce the overall energy consumption of the system.

The remainder of this paper is organized as follows. Section~\ref{sec:bg} provides a quick summary of the neural network models used in this paper, and Section~\ref{sec:NN} presents the methodology used for selecting hyperparameters of the models and for training.
Section~\ref{sec:deviation} then describes the modeling of deviations, that is of the impact of circuit faults on the computations. Section~\ref{sec:results} discusses the robustness of the inference based on simulation results. Finally, Section~\ref{sec:conclusion} concludes the paper.

\section{Background and Naming Conventions}\label{sec:bg}
A neural network is a neural representation of a function $f$ that is a composition of \textit{layer} functions $f_i$. Each layer function is usually composed of a linear operation followed by a non-linear one.
A \textit{dense layer} is such that its inputs and outputs are vectors. Its linear part can be written as a matrix multiplication between an input $x$ and a \textit{weight matrix} $W$: $x \mapsto Wx$.
The \textit{number of neurons} of a dense layer refers to the number of rows of $W$.
A \textit{$n$-dimensional convolutional layer} is such that its inputs and outputs are tensors of rank $n$. Its linear part can be written as a n-dimensional convolution between an input $x$ and a \textit{weight tensor} $W$ of rank $n+2$: $x \mapsto (\sum\limits_p{W_{pq} \ast_n x_p})_{\forall q}$,
where $p$ and $q$ index slices at the last ranks and $\ast_n$ denotes the $n$-dimensional convolution. The output tensor slices indexed by $p$ and $q$ are called \textit{feature maps}.
A pooling operation is often added to convolutional layers to scale them down. 
%Dropout~\cite{srivastava2014dropout} operation is commonly used. It consists in setting a random number of neurons to zero during the training phase to prevent overfitting. This operation is compensated at test time by scaling down the whole layer.

An MLP~\cite{hornik1989multilayer} is composed only of dense layers.
A CNN~\cite{lecun1998gradient} is mainly composed of convolutional layers.
For both network types, to perform supervised classification, a dense output layer with as many neurons as classes is usually added.
Then, the weights are trained according to an optimization algorithm based on gradient descent.
%We measure the error between an output and its expected output with a discriminative loss function $\mathcal{L}$. During the training phase, the weights of the network are adapted for the classification task based on the errors that are back-propagated~\cite{hornik1989multilayer} via the chain rule and according to a chosen optimization algorithm \cite{bottou2010large}.

\section{Neural Network Models}\label{sec:NN}

We consider two types of deep learning models and train them in the usual way, assuming reliable computations.
To simplify model selection, we place some mild restrictions on the hyperparameter space,
since we are more interested in the general robustness of the models, rather than in finding models with the very best accuracy.
As described below, we restrict most layers to have the same number of neurons, and the same activation function. We also try only one optimisation algorithm, and consider only one type of weight initialization.

The first model type that we consider is an MLP network composed of $L$ dense layers, each containing $N$ neurons, that we denote as MLP--$L$--$N$.
The activation function used in all the layers is chosen as the rectified linear unit (reLU)~\cite{glorot2011deep}. 
In fact, since a circuit implementation (and particularly, a fixed-point circuit implementation) can only represent values over a bounded range, we take this into account in the training by using a clipped-reLU activation, which adds a saturation operation on positive outputs. We note that such an activation function has been used before in a different context, in order to avoid the problem of diverging gradients in the training of recurrent networks~\cite{clippedRelu2014}.
The use of the $\tanh$ activation was also considered, but reLU was observed to yield better accuracy.

The second model type is a CNN network composed of $L$ convolutional $C \times C$ layers with $P \times P$ pooling of type $pool$, ultimately followed by a dense layer of 200 neurons~\cite{lecun1998gradient}. This class is denoted as CNN--$L$--$C$--$P$--$F$--$pool$, where $F$ is the number of feature maps used in each convolutional layer.
Clipped-reLU activations are used throughout.
The type of pooling $pool$ can be either ``max'' pooling or no pooling.

For simplicity, we opted to train the networks on the task of digit classification using the MNIST dataset \cite{lecun1998mnist}. In addition to the layers defined above, each model is terminated by a dense ``softmax'' layer used for classification, and containing one neuron per class.
Initialization was done as suggested in \cite{glorot2010understanding}. To prevent overfitting during training, a dropout~\cite{srivastava2014dropout} of 25\% and 50\% of the neurons have been applied on convolutional and dense layers, respectively, except on the first layer. We used categorical crossentropy as the loss function and the ``adadelta'' optimizer \cite{matthew2012adadelta}. The batch size was 128 and we trained for 15 epochs.
The saturation value of the clipped-reLU activation is chosen as $1$ as this provides a good balance between performance and implementation complexity. Note that as the range is increased, more bits must be used to maintain the same precision, leading to larger circuits.

\section{Deviation Models}\label{sec:deviation}

We consider that all the computations performed in the inference phase are unreliable, either because they are performed by unreliable circuits, or because the matrix or tensor $W$ is stored in an unreliable memory. 
We do, however, assume that the softmax operation performed at the end of the classification layer (i.e.~the output layer) is computed reliably.
We assume that each layer is affected by deviations independently, and that deviations occur independently for each scalar element of a layer output.
Note that in the case of a convolutional layer, it would be natural in practice to store each convolution kernel only once, even though each kernel is used to generate multiple output scalar elements. The assumption that deviations occur independently for each scalar output is therefore a simplifying assumption for the case where the tensor $W$ of a convolutional layer is stored unreliably. However, it remains realistic for the case where deviations are caused by the processing circuits.
%Note: It is reasonable to assume that deviations occur independently in each layer since in a circuit implementation, it would be natural to place registers or perform memory storage between layers, with the result that faults occurring in one layer would not influence the occurrence of deviations in other layers. Furthermore, different layers rely on different set of parameters.

We are interested in circuits with a reasonably low amount of unreliability, and therefore it is useful to partition the outcome of a computation in terms of the occurrence or non-occurrence of a deviation event. 
We say that a deviation event occurs if the output of a circuit is different from the output that would be generated by a fully reliable circuit. The probability of a deviation event is denoted by $p$.

Obtaining a precise characterization of the output of a circuit when timing violations or other circuit faults are possible requires knowledge of the specific circuit architecture and of various implementation parameters.
We will rely here on two simplified deviation models. 
We assume that the circuits operate on fixed-point values defined over a certain range, which is motivated by the fact that fixed-point computations are much simpler than floating-point ones, and reducing circuit complexity is the obvious first step in trying to improve energy efficiency.
The first deviation model is a pessimistic model that assumes that when a deviation occurs, the output is sampled uniformly at random from the bounded output domain of that circuit. We call this deviation model \emph{conditionally uniform}. 
Note that we assume that the circuit output is continuous within the bounded range to avoid the need to take into account the number of quantization levels. 
%Note: If instead the circuit output was defined over a finite set, the deviations could be modeled as a $q$-ary symmetric channel.
%Note: The model is pessimistic if we assume that the faulty output is composed of some of the bits of the previous output, and some of the bits of the current output. Since the uniform distribution maximizes the entropy, it will introduce the worst error. (This statement would need to be checked carefully before putting in a paper.)
The second model is more optimistic, and assumes that the occurrence of a deviation can be detected. Therefore, when a deviation occurs, we replace the output by a ``neutral'' value, in this case $0$. We call this deviation model the \emph{erasure} model.

In a synchronous circuit, the deviations can be observed in the memory elements (registers) that separate the logic circuits. By changing the placement of these registers, we can in effect select the point where deviations will be taken into account. Note however that register placement cannot be arbitrary, as we also seek to have similar logic depth separating all registers.
When considering the effect of the deviations on the inference, we noticed that robustness can be increased if deviations are sampled before the activation function of each layer. This is not surprising, since these activation functions act as denoising operations. All simulation results presented in this paper therefore consider that deviations are sampled before the activation function. In the case of convolutional layers, the pooling operation is also performed after the sampling of deviations.

\section{Results}\label{sec:results}

The classification performance of the MLP and CNN models was evaluated using Monte-Carlo (MC) simulations that sample deviations according to the deviation models described in Section~\ref{sec:deviation}.
Because we wish to evaluate a large number of neural network models, each MC simulation is performed by sampling only 10 deviation realizations.

\begin{figure}[tp]
  \centering
  \includegraphics[height=7cm]{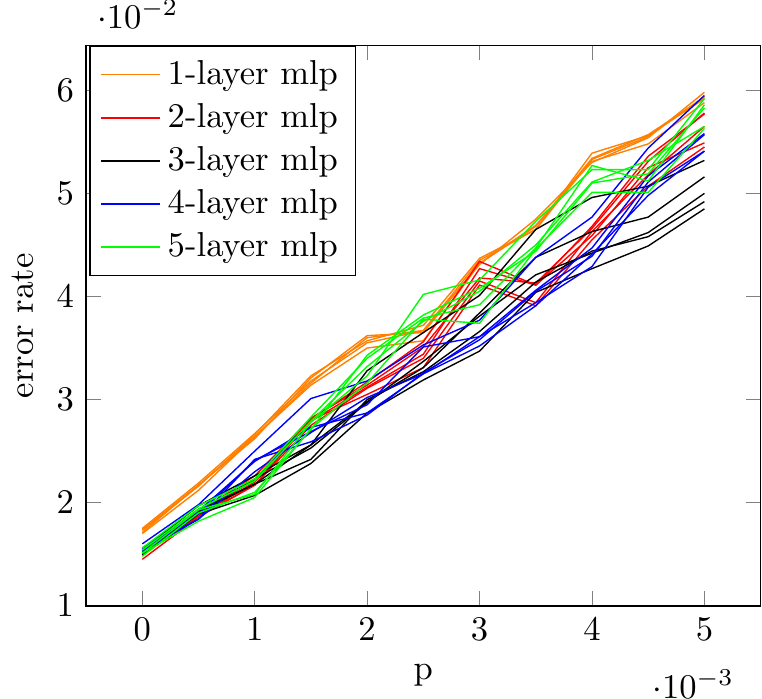}
  \caption{Error rate of MLP under conditionally uniform deviations with prob.~$p$. (Best viewed in color.)}
  \label{fig:r_mlp}
\end{figure}

\begin{figure}[tp]
  \centering
  \includegraphics[height=7cm]{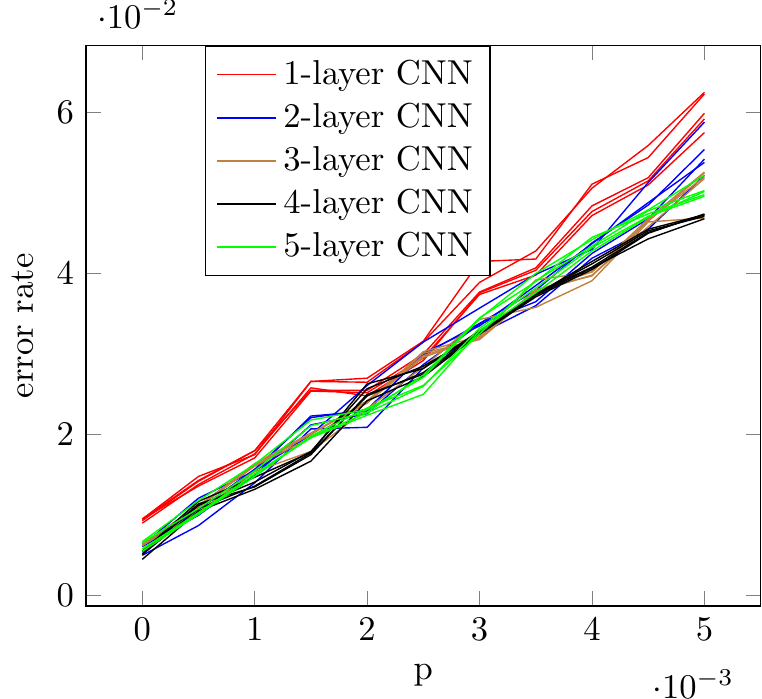}
  \caption{Error rate of CNN under conditionally uniform deviations with prob.~$p$. (Best viewed in color.)}
  \label{fig:r_cnn}
\end{figure}

\subsection{Effect of some hyperparameters on robustness}

Robustness refers to the ability to maintain good classification accuracy in the presence of deviations. A model has better robustness if it achieves a lower classification error at a given $p$.
We investigated the impact of several hyperparameters on the robustness of the inference. We evaluated the performance using the clipped-reLU and $\tanh$ activation functions, and found robustness to be similar in both cases. 
For CNN models, we also evaluated the impact of the choice of pooling function. In this case, we found that using no pooling, rather than max pooling, provided a slight improvement in robustness.
Finally, we considered the impact of the number of layers $L$.
The effect of $L$ on classification error is shown in Figure~\ref{fig:r_mlp} for an MLP-$L$-$N$ network and in Figure~\ref{fig:r_cnn} for a  CNN-$L$-$C$-$P$-$F$-$pool$ network, for the case where the inference is affected by conditionally uniform deviations.
For each value of $L$, we select the 5 best models based on their deviation-free performance to show the variability of the model optimization.
In both cases, the number of layers does not have a clear impact on robustness, but depending on the value of $p$, increasing the number of layers can improve robustness, even when a larger number of layers decreases performance at $p=0$ (reliable computation). For example, in the case of MLP models, the best performance at $p=0$ is obtained by a 2-layer model, but using a larger number of layers improves performance for $p\geq 5 \cdot 10^{-4}$. %Note: 5 \cdot 10^{-4} is the first measured point for which this is true.

\subsection{Fault tolerance}
% Here we show how equivalent performance can be obtained using faulty hardware, at the cost of an increase in the number of parameters.
% We could also discuss cases where we cannot find a faulty network that matches the reliable-hardware performance
Since neural networks are designed to reject noise in their input, we might expect them to also be able to reject the ``noise'' introduced by faulty computations. We investigate this ability for networks trained with standard procedures, in order to provide a baseline for future targeted approaches.

\begin{figure}[tp]
  \centering
  \includegraphics[height=6.8cm]{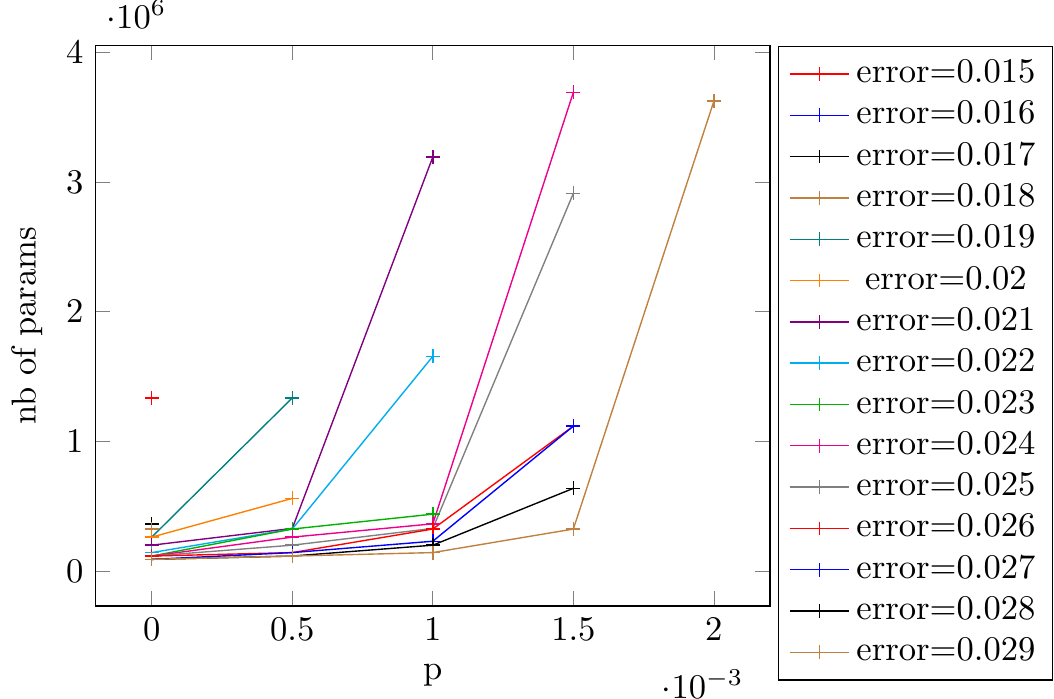}
  \caption{Number of model parameters needed to achieve error rate target using MLP under conditionally uniform deviations.}
  \label{fig:mlp_nvsp_conduni}
\end{figure}

\begin{figure}[tp]
  \centering
  \includegraphics[height=6.8cm]{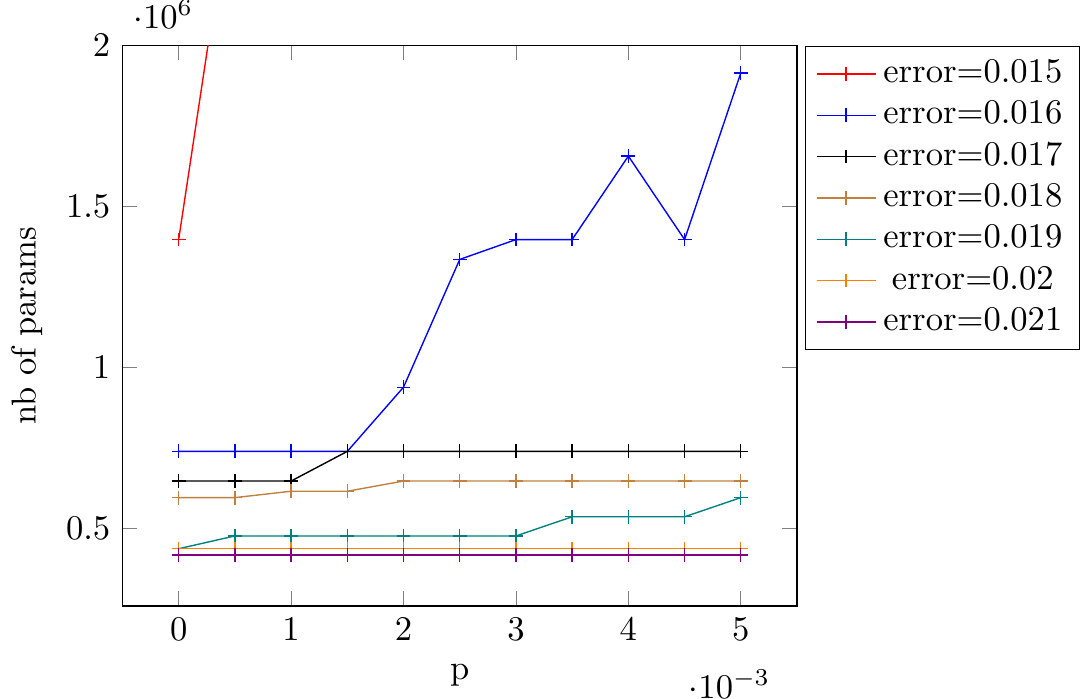}
  \caption{Number of model parameters needed to achieve error rate target using MLP under erasure deviations.}
  \label{fig:mlp_nvsp_erasure}
\end{figure}

\begin{figure}[tp]
  \centering
  \includegraphics[height=6.8cm]{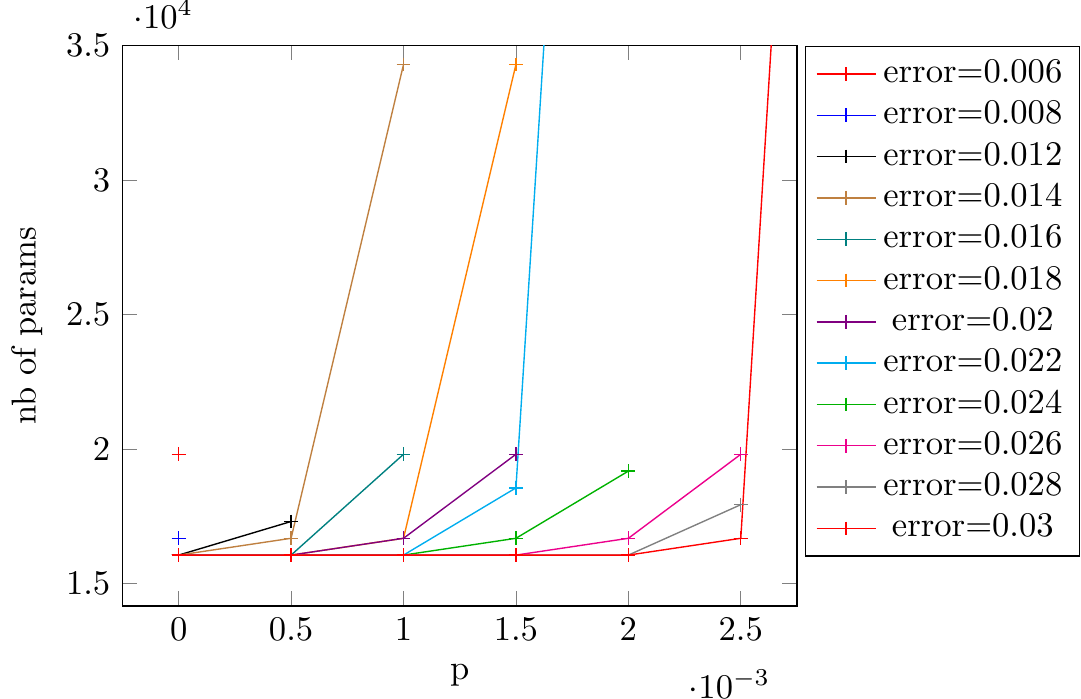}
  \caption{Number of model parameters needed to achieve error rate target using CNN under conditionally uniform deviations.}
  \label{fig:cnn_nvsp_conduni}
\end{figure}

\begin{figure}[tp]
  \centering
  \includegraphics[height=6.8cm]{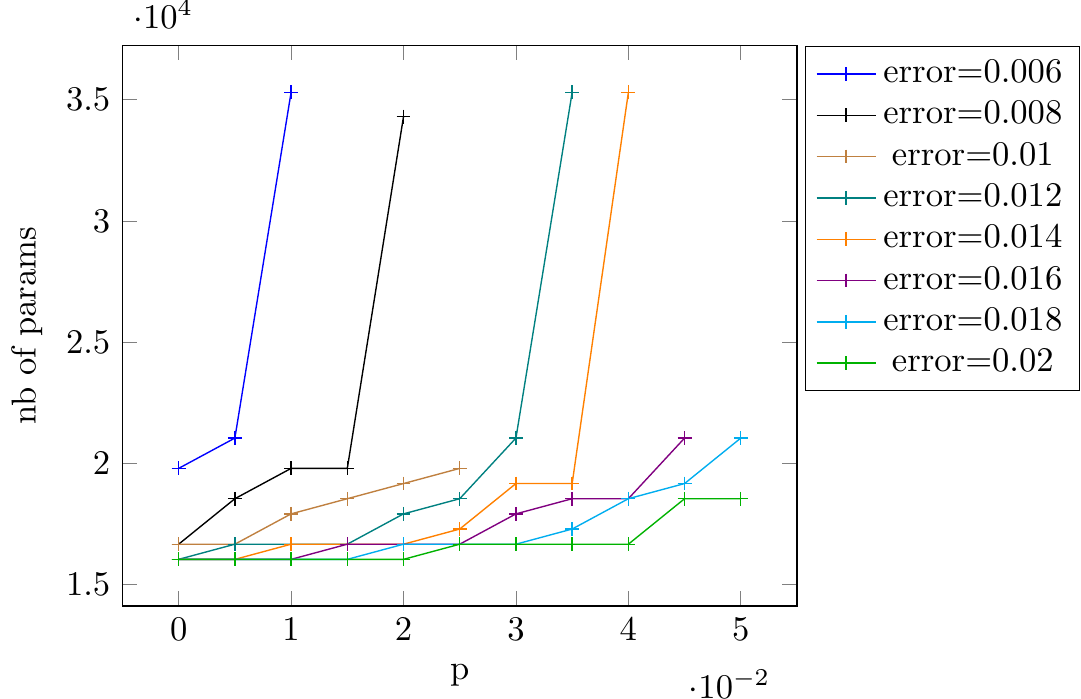}
  \caption{Number of model parameters needed to achieve error rate target using CNN under erasure deviations.}
  \label{fig:cnn_nvsp_erasure}
\end{figure}

We first choose an error rate target that we want the network to achieve. We then consider various deviation probability values, and for each, look for the model with the smallest number of parameters that can achieve or outperform the performance target under the deviation constraint.
For MLP networks, the results are shown in Figure~\ref{fig:mlp_nvsp_conduni} for the case of conditionally uniform deviations, and in Figure~\ref{fig:mlp_nvsp_erasure} for the case of erasure deviations. Similarly, Figures~\ref{fig:cnn_nvsp_conduni} and \ref{fig:cnn_nvsp_erasure} show the results for CNN networks, respectively for conditionally uniform and erasure deviations.

Note that the finite set of models trained implies upper and lower bounds on the number of parameters.
As a result, the lack of robust models for some performance targets (represented by single data points at $p=0$) does not necessarily mean that deviations cannot be tolerated at that performance target, but merely that no sufficiently large model were found within the set of models that were trained. Similarly, the saturation to a minimum number of parameters that is observed as the performance target is decreased corresponds to the number of parameters in the smallest trained models.

We simulate deviation probabilities on the order of $10^{-3}$, which are already considered quite large for digital circuit design. 
We can see that for both network types and for both deviation models, there are indeed many performance targets at which performance can be preserved in the presence of computation failures by picking a larger model from the set of pre-trained models.
% the necessary redundancy grows slower for the erasure model
As might be expected, the necessary redundancy grows more slowly for the optimistic erasure deviation model. In fact for this deviation model, CNN models are able to preserve performance even for deviation probabilities on the order of $10^{-2}$, as seen in Fig.~\ref{fig:cnn_nvsp_erasure}.

% explain apparent inconsistencies
% zigzag in fig. 4, smaller number of parameters required for larger p: probably because only 10 deviation realizations are considered in the simulations (this allows simulating a large number of models)
We expect all the curves in Figures~\ref{fig:mlp_nvsp_conduni}--\ref{fig:cnn_nvsp_erasure} to be non-decreasing, since computation noise is unlikely to improve performance in a feedforward network.
The apparent inconsistency observed in Fig.~\ref{fig:mlp_nvsp_erasure} for the $0.016$ error target is likely an artefact caused by the small number of deviations that are sampled in the MC simulation.

% Efficiency discussion
% introduce and define efficiency: nb of params at p=0 / nb of params
In addition to considering the absolute number of parameters required to achieve a performance target in the presence of deviations, it is interesting to compare the number of parameters required by the faulty and reliable implementations, to obtain the fault-tolerance efficiency of the network. 
Let $M$ be a network model that achieves our performance target when deviations occur with probability $p$, and $M_o$ be the smallest model that achieves the target under reliable computation. We define the fault-tolerance efficiency of $M$ as $n(M_o) / n(M)$, where $n(X)$ denotes the number of parameters used by a model $X$.
To evaluate efficiency empirically, we take $M_o$ to be the smallest available model that achieves the performance target, even though it is possible that a smaller model exists but was not found during the training phase.

\begin{figure}[tp]
  \centering
  \includegraphics[height=7cm]{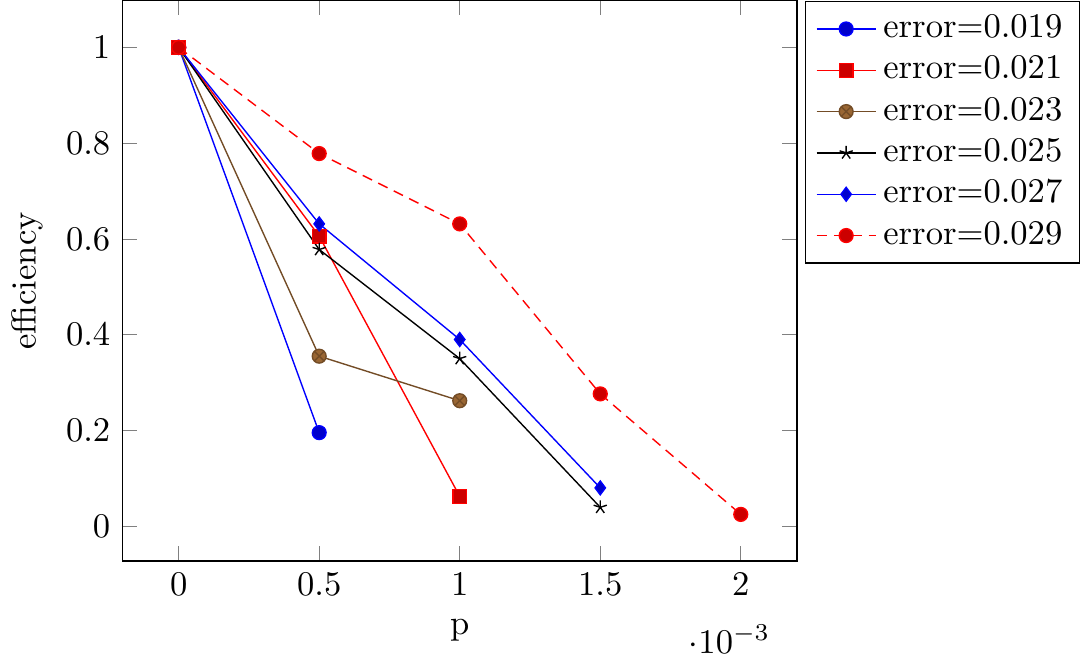}
  \caption{Fault-tolerance efficiency under an error rate target (MLP, conditionally uniform deviations).}
  \label{fig:mlp_nvsp_conduni_bis}
\end{figure}

\begin{figure}[tp]
  \centering
  \includegraphics[height=7cm]{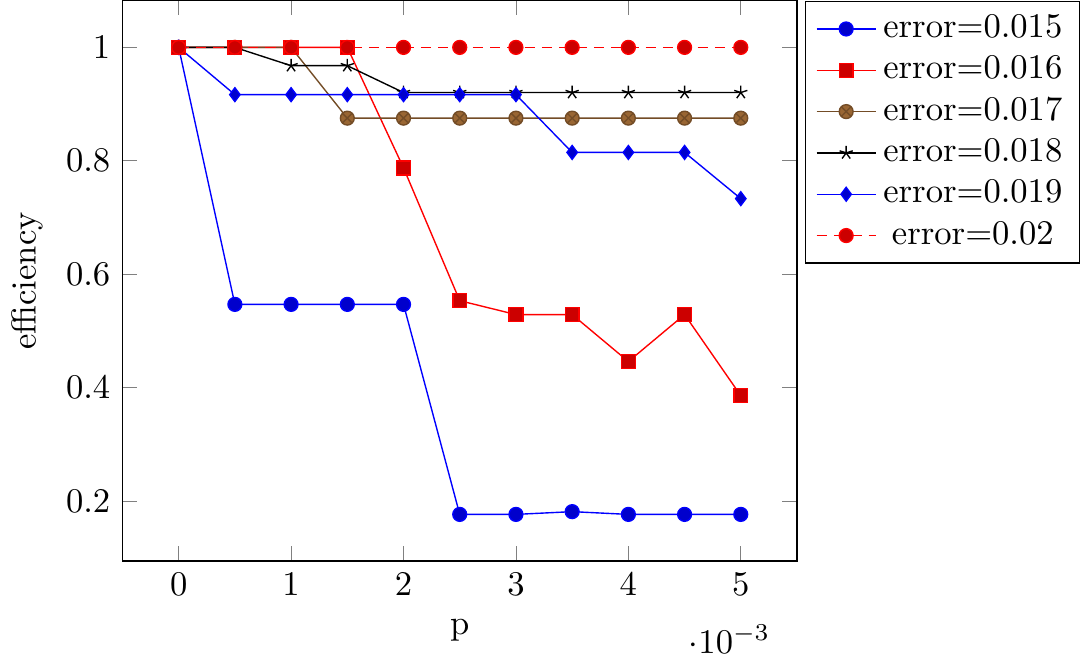}
  \caption{Fault-tolerance efficiency under an error rate target (MLP, erasure deviations).}
  \label{fig:mlp_nvsp_erasure_bis}
\end{figure}

\begin{figure}[tp]
  \centering
  \includegraphics[height=7cm]{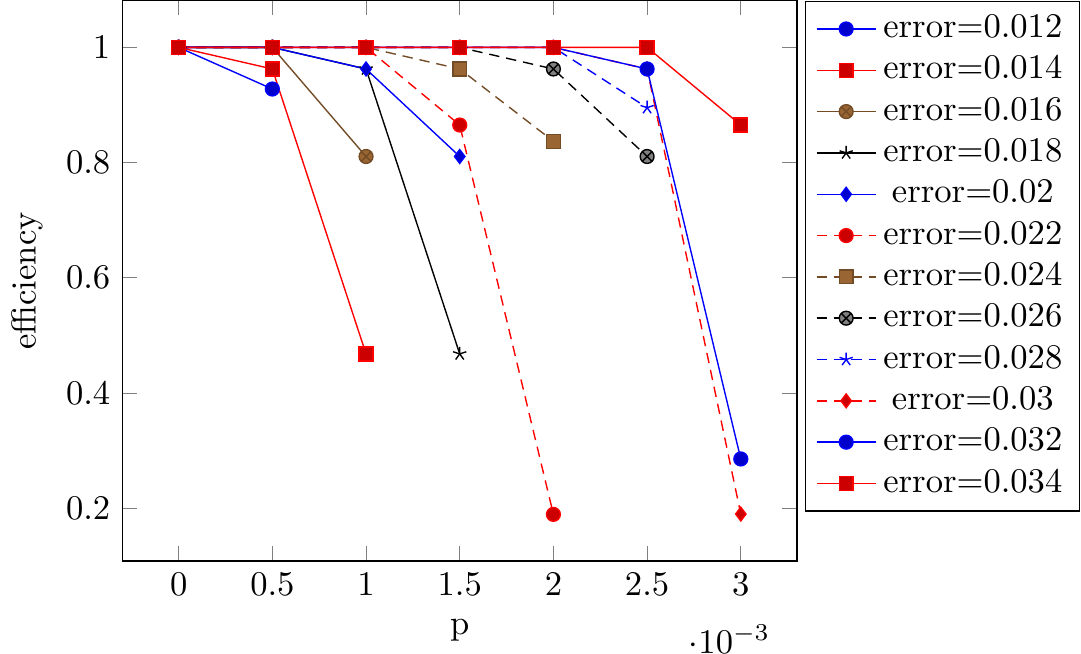}
  \caption{Fault-tolerance efficiency under an error rate target (CNN, conditionally uniform deviations).}
  \label{fig:cnn_nvsp_conduni_bis}
\end{figure}

\begin{figure}[tp]
  \centering
  \includegraphics[height=7cm]{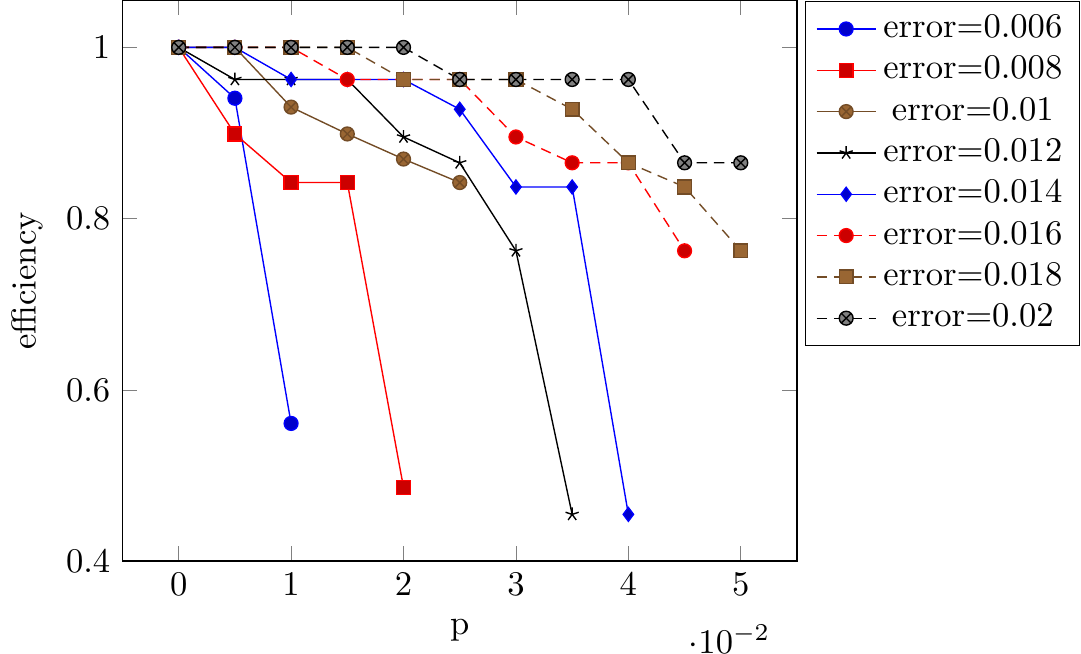}
  \caption{Fault-tolerance efficiency under an error rate target (CNN, erasure deviations).}
  \label{fig:cnn_nvsp_erasure_bis}
\end{figure}

Efficiency curves for the MLP models are shown in Figures~\ref{fig:mlp_nvsp_conduni_bis} and \ref{fig:mlp_nvsp_erasure_bis}, and for the CNN models in Figures~\ref{fig:cnn_nvsp_conduni_bis} and \ref{fig:cnn_nvsp_erasure_bis}.
Two characteristics of these results are worth pointing out.
First, even if, in the case of CNN models, a significant amount of deviations can be tolerated while maintaining an efficiency close to 1, there exist clear deviation thresholds after which the efficiency starts degrading rapidly.
Second, this deviation threshold becomes smaller as the error rate target is decreased.
This suggests that with this naive fault-tolerance approach, we are forced to compromise on fault-tolerance efficiency to achieve a lower classification error rate.
%Therefore, better fault-tolerance approaches are needed in order to remove this dependency of the efficiency on the classification performance.

% - CNN models can tolerate a significant amount of deviations while maintaining a good efficiency
% - but the results show a clear threshold at which efficiency starts degrading rapidly
% - and this threshold becomes smaller as the error rate target is decreased
% This suggests that with this naive fault-tolerance approach, we are forced to compromise on fault-tolerance if we want to achieve a lower classification error rate.
% Intuitively, we would think that it is possible to find fault tolerance scheme for which the efficiency is independent of the error rate target.

% Similar comments apply to the MLP case, except that the fault tolerance is not as good (even if we allow about twice the error rate compared to the best model found, the efficiency is worse than a CNN with doubled error rate w.r.t best model)

% occasionally some curves show better efficiency for a better error rate: also due to the small number of deviation realizations considered in the simulations and/or the uncertainty regarding n(M_o).

\section{Conclusion}\label{sec:conclusion}

In this paper, we have considered the effect of faulty computations on the performance of MLP and CNN inference, using a pessimistic and an optimistic deviation model.
We showed that using standard training procedures, it is in many cases possible to find models that will compensate for computation failures, when the deviation probability is on the order of $10^{-3}$ and at the cost of an increase in the number of parameters. 
% Results show that efficiency is linked to classification performance.
% Of course, efficiency is inevitably linked with the deviation probability, but it is reasonable to expect that networks designed for fault-tolerance could achieve an efficiency that does not depend on the performance target.
We also studied the efficiency of the fault-tolerance that is achieved using standard model training. Our results show that the fault-tolerance efficiency decreases as the performance target is increased.
It seems reasonable to expect that networks designed for fault-tolerance could achieve an efficiency that only depends on the amount of deviations occurring during the inference process, and not on the performance target.
These results therefore provide a baseline for future work seeking to identify systematic ways of designing robust deep neural networks, and highlight that one of the objectives of the network design should be to decouple the fault tolerance from the classification performance.

%%%%%%%%%%%%%%%%%%%%%%%
\bibliography{IEEEabrv,cognitive17.bib,bunch.bib}
\bibliographystyle{IEEEtran}

\end{document}